\documentclass[runningheads]{llncs}
\usepackage[T1]{fontenc}
\usepackage{graphicx}
\usepackage{booktabs}
\usepackage[misc]{ifsym}

\usepackage{mwe}
\usepackage{bbm}
\usepackage{amsfonts}
\usepackage{makecell}
\newcommand{\ie}{i.e.}
\newcommand{\etal}{et al.}
\newcommand{\eg}{e.g.}
\newcommand{\softmax}{\mbox{softmax}}
\usepackage{multirow}

\begin{document}

\title{AKGNet: Attribute Knowledge-Guided Unsupervised Lung-Infected Area Segmentation}

\titlerunning{AKGNet}
\author{Qing En \and Yuhong Guo}
\authorrunning{Q. En, Y. Guo}

\institute{School of Computer Science, Carleton University, Ottawa, Canada}

\maketitle              

\begin{abstract}
Lung-infected area segmentation is crucial for assessing the severity of lung diseases. 
However, existing image-text multi-modal methods typically rely on labour-intensive annotations for model training, posing challenges regarding time and expertise. 
To address this issue, we propose a novel attribute knowledge-guided framework for unsupervised lung-infected area segmentation (AKGNet), which achieves segmentation solely based on image-text data without any mask annotation.
AKGNet facilitates text attribute knowledge learning, attribute-image cross-attention fusion, and high-confidence-based pseudo-label exploration simultaneously. 
It can learn statistical information and capture spatial correlations between image and text attributes in the embedding space, iteratively refining the mask to enhance segmentation.
Specifically, we introduce a text attribute knowledge learning module by extracting attribute knowledge and incorporating it into feature representations, enabling the model to learn statistical information and adapt to different attributes. 
Moreover, we devise an attribute-image cross-attention module by calculating the correlation between attributes and images in the embedding space to capture spatial dependency information, thus selectively focusing on relevant regions while filtering irrelevant areas. 
Finally, a self-training mask improvement process is employed by generating pseudo-labels using high-confidence predictions to iteratively enhance the mask and segmentation.
Experimental results on a benchmark medical image dataset demonstrate the superior performance of our method compared to state-of-the-art segmentation techniques in unsupervised scenarios.

\end{abstract}

\section{Introduction}
Medical image analysis is crucial for diagnosing lung diseases, particularly pneumonia and tuberculosis \cite{duncan2000medical,shen2017deep,anwar2018medical}. 
A fundamental task within this domain is lung-infected region segmentation, aimed at identifying infected regions in medical images \cite{mansoor2015segmentation,fan2020inf}. 
With the advancements in deep neural networks, fully-supervised medical image segmentation methods have emerged, often relying on densely-labeled masks for model training \cite{ronneberger2015u,chen2021transunet}. 
Additionally, some approaches utilize image-text pairs to develop more expressive image-text models, thus enhancing segmentation performance in the medical domain \cite{li2023lvit}.

While existing fully supervised methods have shown progress, they typically depend on manual annotation for model training, which involves meticulous delineation of infected regions by domain experts \cite{tajbakhsh2020embracing}. 
In contrast, image-text unsupervised lung-infected area segmentation methods can achieve segmentation without explicit manual annotation by autonomously learning feature representations from existing medical images and text data, thereby capturing implicit relationships. 
This paper aims to investigate an untouched task: image-text 
based
unsupervised lung-infected area segmentation, which can effectively achieve lung-infected area segmentation without mask annotations.
However, the absence of mask guidance makes the task not easy. 
It is inappropriate to directly train existing fully-supervised models with
automated
coarse masks as training targets, as they ignore valuable text attribute knowledge in textual information and 
enhance noise from the coarse masks. 

\begin{figure}[t]
\centering
  \includegraphics[width=0.95\linewidth,height=38mm]{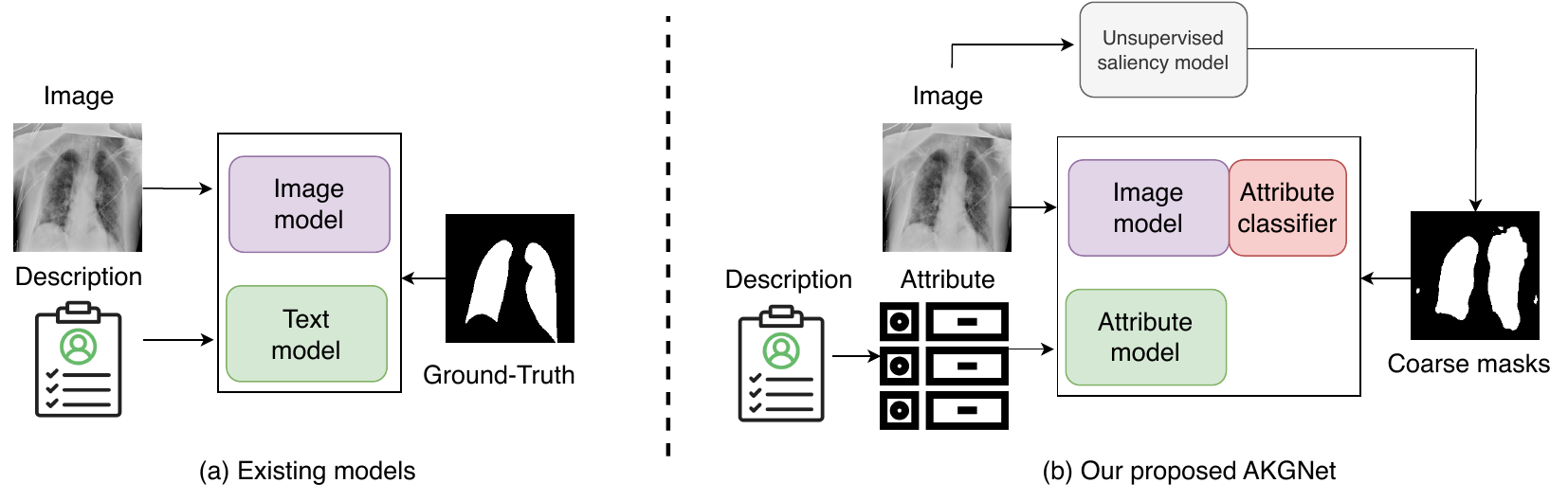}\\
\caption{
Illustration of the proposed idea.
(a) Existing methods require mask annotations to train the model to achieve image-text lung infection region segmentation. 
(b) Our proposed method does not require mask annotation to achieve image-text unsupervised lung infection region segmentation by mining the valuable text attribute knowledge to learn statistical information.
}
  \label{fig:idea}
\end{figure}

The challenges in image-text 
based
unsupervised lung-infected area segmentation can be summarized in two aspects:
(1) The absence of mask annotation deprives the model of a direct and reliable training target, which is particularly challenging for complex heterogeneous lung infections.
(2) Redundant information in textual descriptions may hinder model training, and integrating textual and image information unsupervisedly introduces additional difficulty.
Drawing from human cognitive science, humans can extract regularities from the environment over time \cite{sherman2020prevalence}.
We observe that text descriptions contain valuable text attribute knowledge, significantly enhancing the understanding and interpretation of visual data. 
This text attribute knowledge encompasses critical details such as pathological findings and clinical observations, which can provide additional statistical information for model learning.
Motivated by this observation, we aim to address the challenges above by extracting and leveraging valuable text attribute knowledge from textual descriptions to enhance segmentation efficacy.

In this paper, we propose a novel framework, AKGNet, for segmenting lung-infected areas based on image-text data without mask annotation, as shown in Fig \ref{fig:idea}.
AKGNet can leverage text attribute knowledge to learn statistical information, facilitate attribute-image cross-attention to capture spatial correlations between image and text attributes, and iteratively refine masks by exploring high-confidence-based pseudo-labels.
Specifically, we introduce a text attribute knowledge learning module that extracts attribute knowledge and integrates it into feature representations. 
This allows the model to acquire statistical information and adapt to various attributes by calculating the proposed attribute classification loss using mask-guided features and attribute targets.
Moreover, we develop an attribute-image cross-attention module that calculates correlations between attributes and images in the embedding space. This captures spatial dependency information, enabling selective focus on relevant regions while filtering out irrelevant areas.
Furthermore, a self-training mask refinement process is utilized to improve the mask by generating pseudo-labels from high-confidence predictions. 
This iterative approach enhances the mask and segmentation results by calculating the proposed self-training loss.
The unsupervised segmentation loss, derived from generated coarse masks, is integrated with the previously mentioned losses to jointly optimize AKGNet.
The main contributions of our paper can be summarized as follows:
\begin{itemize}
\item 
We propose a novel AKGNet for unsupervised lung-infected area segmentation based on image-text data without mask annotation.

\item 
AKGNet enables simultaneous learning of text attribute knowledge, cross-attention fusion between attributes and images, and exploration of high-confidence pseudo-labels. It efficiently captures statistical information and spatial correlations between image and text attributes in the embedding space, iteratively refining the mask to improve segmentation.

\item 
Experimental results on a benchmark medical image dataset show that the proposed AKGNet can effectively perform unsupervised lung-infected area segmentation.

\end{itemize}

\section{Related Work}
\subsection{Medical Image Segmentation}
Medical image segmentation aims to assign specific labels to each pixel within an image, encompassing organ classification and lesion area delineation. 
Existing segmentation models typically fall into two categories: Convolutional Neural Network (CNN) and Transformer architectures \cite{milletari2016v,ronneberger2015u,chen2021transunet,wang2022mixed,li2022tfcns}.
UNet \cite{ronneberger2015u} is a widely used CNN-based model known for its efficient encoder-decoder structure and effective segmentation results. 
Additionally, models incorporating attention mechanisms \cite{zhang2019net,fan2020pranet,rahman2023medical} and denser connections \cite{zhou2018unet++} based on UNet have shown promise in improving medical image segmentation. 
Inspired by their success in natural scene segmentation, transformer-based methods employ global attention mechanisms like self-attention and trans-attention to capture medical image characteristics \cite{dong2021polyp,cao2023swin,wang2022mixed}. 
Some methods combine CNN and transformer structures to leverage global and local semantic information \cite{chen2021transunet,li2022tfcns}. 
While previous approaches focus solely on image data and require dense labeling, our work concentrates on image-text unsupervised lung-infected area segmentation.

\subsection{Vision-language Segmentation}
Recently, vision-language models have made significant strides in multi-modal visual recognition tasks \cite{radford2021learning,zhang2023large}. 
Notably, the CLIP model \cite{radford2021learning} stands out, utilizing extensive text-image pairs to train its transformer-based image and text encoders. 
This training process maximizes the similarity between positive image and text embeddings while minimizing the similarity of embeddings from negative pairs. 
Moreover, various vision-language models have been proposed to enhance natural and medical image segmentation by incorporating text descriptions as prompt information \cite{poudel2023exploring}. 
Referring image segmentation methods add decoder on top of the pre-trained image and text encoders to integrate vision and language information, facilitating segmentation \cite{luddecke2022image,wang2022cris}. 
Similarly, image-text medical image segmentation methods utilize textual descriptions as a reference to fuse visual and linguistic information through hybrid CNN and transformer architectures \cite{li2023lvit,lee2023text}. 
However, these methods rely on masked labeled information and do not fully exploit valuable textual descriptions. 
In contrast, our proposed framework achieves image-text medical image segmentation without requiring mask annotation, leveraging text attribute information mined from text descriptions.

\section{Method}
This section introduces the AKGNet framework for unsupervised segmentation of lung-infected areas. 
We provide an overview of the architecture of AKGNet in Section \ref{sec:overview}, followed by the coarse mask generation process detailed in Section \ref{sec:CMG}. 
Next, we describe the proposed text attribute knowledge learning module in Section \ref{sec:AKLM} and the attribute-image cross-attention module in Section \ref{sec:AICAM}. 
Then, we present the self-training mask refining process in Section \ref{sec:STMR}. 
Finally, we outline the overall loss function utilized for training AKGNet in Section \ref{sec:loss}.

\begin{figure}
\centering
  \includegraphics[width=0.88\linewidth,height=50mm]{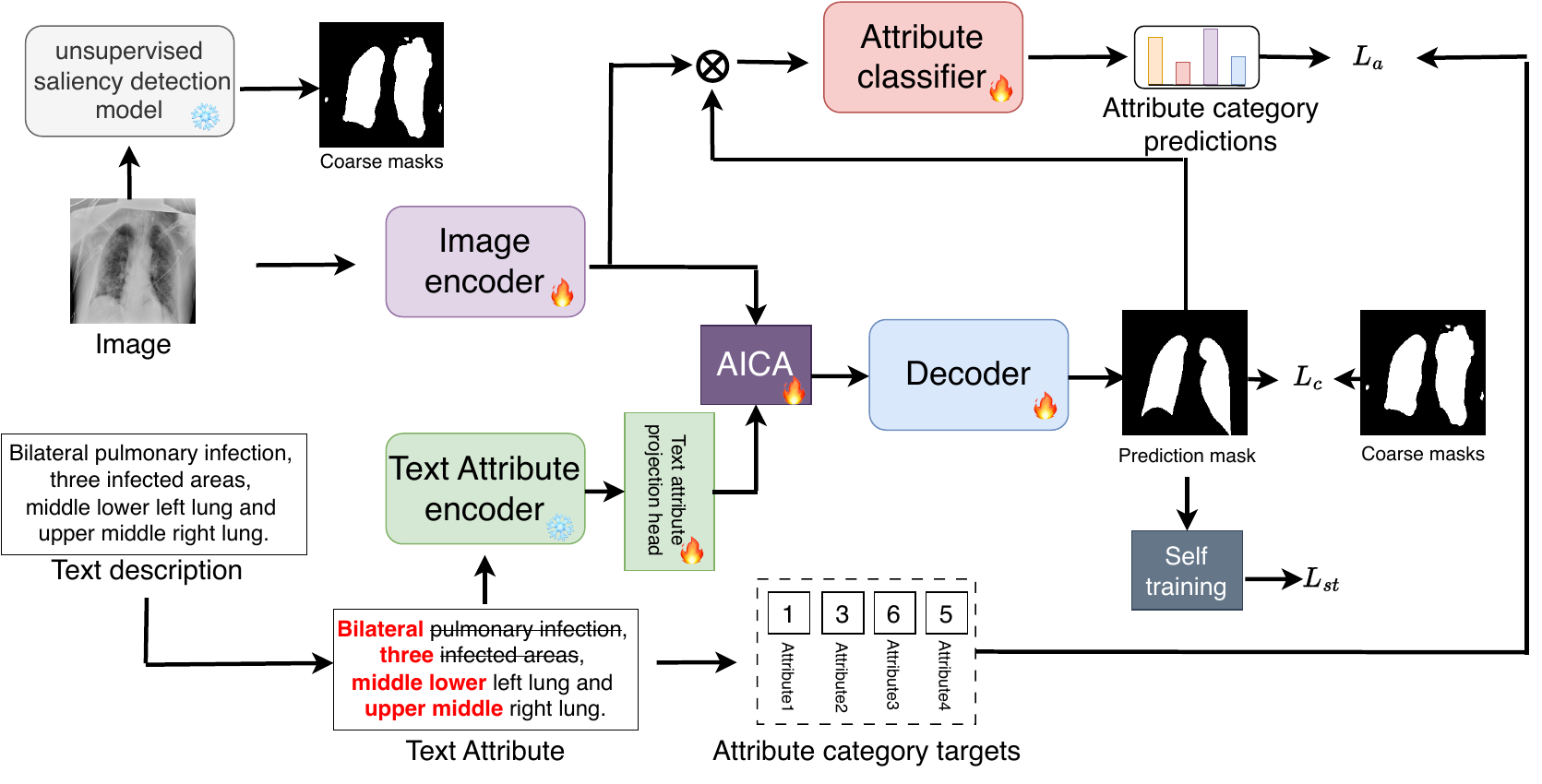}\\
\caption{
An overview of the proposed AKGNet. 
First, a coarse mask is generated. 
Next, text attribute knowledge is extracted from text descriptions to construct the training target for the text attribute classifier. 
Then, the mask-guided image features extracted from the image encoder are fed into the classifier to compute $L_{a}$. 
The attribute-image fusion features generated by AICA module are fed into the image decoder to generate a prediction mask, which is used to compute $L_{c}$ with the coarse mask.
Finally, the self-training mask refining process is implemented by computing $L_{st}$.
}
  \label{fig:framework}
\end{figure}

\subsection{Overall Framework}
\label{sec:overview}
In the unsupervised segmentation of lung-infected areas, our objective is to train a robust segmentation model using a limited set of $N$ image-text pairs, denoted as $\mathcal{D}=\{(I^{n},T^{n})\}_{n=1}^{N}$. 
Here, $I \in \mathbb{R}^{1\times H\times W}$ represents an input image, and $T\in \mathbb{R}^{L}$ denotes an input textual description, with $H$ and $W$ denoting the height and width of the input image, respectively, and $L$ representing the length of the input textual description. 
The overall architecture of the proposed AKGNet is illustrated in Fig. \ref{fig:framework}.

AKGNet comprises an image encoder $f_{I}: \mathbb{R}^{1\times H\times W} \to \mathbb{R}^{c\times h\times w}$, 
$M$ attribute classifiers $\{e_{m}: \mathbb{R}^{c\times h\times w} \to \mathbb{R}^{a_{m}}\}_{m=1}^{M}$, 
an text attribute encoder $f_{A}: \mathbb{R}^{L} \to \mathbb{R}^{d\times L}$,
an text attribute projection head $f_{proj}: \mathbb{R}^{d\times L} \to \mathbb{R}^{c\times L}$,
and an image decoder $g_{I}: \mathbb{R}^{c\times h\times w} \to \mathbb{R}^{1\times H\times W}$.
Here, $c$, $h$, and $w$ denote the channels, height, and width of feature embeddings, respectively. 
Additionally, $d$ represent the dimension of attribute embedding, while $M$ denotes the number of attributes, $m$ denotes the m-th auxiliary classifier, and $a_{m}$ denotes the number of categories for the m-th auxiliary classifier. 
The image embedding and attribute embedding are denoted as $x_{I}=f_{I}(I)$ and $x_{A}=f_{A}(A)$, respectively. 
The decoder’s output is used to produce the segmentation masks $P$.

AKGNet works 
by first generating coarse masks of lung-infected areas by using an unsupervised lung saliency detection model, providing estimations likely belonging to the infected region. 
Next, the text attribute knowledge learning (TAKL) module is employed to acquire statistical information and adapt to various attributes by extracting attribute knowledge and integrating it into feature representations.
Then, the attribute-image cross-attention module is used to capture spatial dependency information by calculating correlations between attributes and images in the embedding space.
Furthermore, a self-training process is utilized to improve the mask by generating pseudo-labels from high-confidence predictions.

\subsection{Coarse Mask Generation}
\label{sec:CMG}
Given that the infected region typically resides within the interior of the lungs, we generate a coarse mask that specifically targets the lung region. This provides an unsupervised indication of the potential presence of the infected area.
Specifically, given an input image $I$, we utilize an unsupervised lung saliency detection model $N_{sal}$ \cite{de2023deep}
to produce a coarse mask $\hat{Y} \in \mathbb{R}^{1\times H\times W}$ as follows:
\begin{equation}
\hat{Y}=\mathbbm{1}[\sigma(N_{sal}(I))>\tau],
\label{coarse mask}
\end{equation}
Here, $\tau$ denotes a predefined threshold for generating a binary coarse mask; $\sigma$ represents the sigmoid function, and $\mathbbm{1}[\cdot]$ denotes the indicator function.
As a result, the dataset $\mathcal{D}$ is expanded to $\mathcal{D}={(I^{n},T^{n},\hat{Y}^{n}) }_{n=1}^{N}$, containing image-text pairs as inputs along with a coarse mask serving as the coarse segmentation target.

\subsection{Text Attribute Knowledge Learning Module}
\label{sec:AKLM}
Although the coarse mask estimates potential infection areas, it lacks precise supervisory information and may contain noise.
Despite the availability of fully-supervised language-driven medical image segmentation methods that leverage textual descriptions to enhance segmentation through multi-modal learning, they often underutilize the valuable text attribute information within these descriptions.
Therefore, we introduce a Text Attribute Knowledge Learning (TAKL) module to extract text attribute knowledge from textual descriptions and perform attribute classification. It computes an attribute classification loss leveraging mask-guided image features and attribute classification targets. 
This module effectively integrates attribute knowledge into the feature representation, facilitating the learning of statistical information.

\subsubsection{Attribute knowledge extraction}
Initially, we extract the text attribute $A$ from the textual description $T$. 
This attribute information is then utilized to establish attribute category for various attributes, which serves as the training target for the attribute classifiers.
Specifically, for a textual description \textit{`Bilateral pulmonary infection, three infected areas, middle lower left lung and upper middle right lung.'}, we first split it into three parts according to commas to obtain $\{$\textit{`Bilateral pulmonary infection',`three infected areas',`middle lower left lung and upper middle right lung'}$\}$.
We then extract four attributes from the description: unilateral/bilateral infection, number of infected areas, and the location (left or right) of the infection within the lung. 
These text attributes, along with their IDs and values, are summarized in Table \ref{tab:attributes}.
For each textual description $T$, we use its four attribute values to construct corresponding attribute description $A$ (\eg \textit{`Bilateral pulmonary infection, three infected areas, middle lower left lung and upper middle lower right lung.'} $\to$ \textit{`Bilateral, three, middle lower, upper middle.'}).
We construct $M$ attribute category targets $\{C_{m}\}_{m=1}^{M}$ based on the attributes for each sample, thereby enriching each sample with attribute classification information.
Consequently, the augmented dataset comprises attribute category information $\mathcal{D}={(I^{n},\hat{Y}^{n},A^{n},\{C^{n}_{m}\}_{m=1}^{M}) }_{n=1}^{N}$, wherein $A$ supplants $T$, offering a more nuanced representation.
Here, $C^{n}_{m}$ denotes the m-th attribute category corresponding to the n-th sample.

\begin{table}[t]
\renewcommand\arraystretch{1.3}
\caption{\label{tab:attributes}
Attribute knowledge from the Qata-COV19 dataset.
Attribute ID: the ID of the attribute (M=4 in total).
Text Attribute Descriptions: the meaning of the attribute.
Text Attribute Values: the set of constituent elements of the attribute.
Different attribute IDs are used to construct different classifiers, while different attribute values are used to construct the categories of the corresponding classifiers.
}
\begin{center}\small
\setlength{\tabcolsep}{0.8mm}
\begin{tabular}{c|c|c}
\hline
Attribute ID&Text Attribute Descriptions&Text Attribute Values \\
\hline
m=1&unilateral or bilateral of lung infection&unilateral, bilateral\\
\hline
m=2&number of infected areas&one,two,three,four,five,six\\
\hline
m=3&\makecell[c]{location of the infected area\\ in the left part} & \makecell[c]{all,upper,middle,lower, \\upper middle, middle lower,no}\\
\hline
m=4&\makecell[c]{location of the infected area\\ in the right part} & \makecell[c]{all,upper,middle,lower, \\upper middle, middle lower,no}\\
\hline
\end{tabular}
\end{center}
\end{table}

\subsubsection{Mask-guided attribute knowledge classification}
After obtaining the attribute category labels, a conventional approach involves classifying the intermediate features of the segmentation model. However, this method solely focuses on the image encoder, overlooking the image decoder responsible for mask generation. Hence, we introduce mask-guided attribute knowledge classification. This technique utilizes the generated prediction masks to isolate foreground regions within intermediate features, which are then employed in an auxiliary classifier for attribute knowledge classification.

Specifically, given an input image $I$ and a text attribute $A$, we initially generate the image embedding $x_{I} \in \mathbb{R}^{c\times h\times w}$ and the prediction mask $P \in \mathbb{R}^{1\times H\times W}$ as previously described.
Next, we produce the masked image feature embedding $x_{MI} \in \mathbb{R}^{c\times h\times w}$, which encapsulates features pertinent to the foreground of the input image, as follows:
\begin{equation}
x_{MI}=x_{I}\mathbbm{1}[\sigma(P)>\alpha].
\label{attribute classification loss}
\end{equation}
Hence, the attribute classification loss is defined as follows:
\begin{equation}
L_{a}=\sum_{m=1}^{M}L_{ce}(e_{m}(x_{MI}),C_{m}),
\label{attribute classification loss}
\end{equation}
where $L_{ce}$ represents the cross-entropy loss function
and 
$\{e_{m}\}_{m=1}^{M}$ denotes $M$ attribute classifiers.

This module adeptly extracts valuable attribute knowledge from textual descriptions, leading to more detailed descriptions. Consequently, models can seamlessly integrate attribute knowledge into feature representations. 
Moreover, the proposed $L_{a}$ leverages prediction masks and intermediate features, fostering the segmentation model's acquisition of statistical information directly pertinent to the task, thus enhancing the model's ability to adapt to diverse attributes.

\subsection{Attribute-Image Cross-Attention Module}
\label{sec:AICAM}
In unsupervised lung infection region segmentation, it is critical to effectively exploit image and attribute correlation to enable the model to focus on relevant regions and ignore irrelevant regions  \cite{fu2019dual}.
Hence, we introduce the attribute-image cross-attention (AICA) module to compute correlations between attributes and images in the embedding space.
It can effectively capture spatial dependency information, allowing selective focus on relevant regions while filtering out irrelevant areas.

The AICA module accepts the image embedding $x_{I}$ and attribute embedding $x_{A}$ as input, generating the attribute-image fusion feature $x_{AI}\in \mathbb{R}^{c\times h\times w}$ as output.
Initially, we pass the attribute embedding $x_{A}$ through the text attribute projection head $f_{proj}$ to align the features of dimension $d$ with the channel dimension $c$ of the image features.
We utilize the learnable parameter $\gamma$ to 
yield the projected attribute embedding $x_{proA}\in \mathbb{R}^{c\times h\times w}$ as follows:
\begin{equation}
x_{proA}=Reshape(f_{proj}(x_{A})\times \gamma).
\label{attribute embedding}
\end{equation}
$Reshape$ represents the reshape operation to get the features of dimension $c\times h\times w$.
$\gamma \in \mathbb{R}^{L\times (hw)}$ represents a learnable parameter that aligns the output of $f_{proj}$ to the same dimension as $x_{I}$. 

Afterwards, we calculate the pairwise dot product between the transformed image embeddings and attribute embeddings to obtain the spatial attention map $S \in \mathbb{R}^{hw\times hw}$:
\begin{equation}
S = \softmax(\phi(x_{I})^{T}\theta(x_{proA})),
\label{attention scores}
\end{equation}
where $\softmax$ represent the softmax function;
$\phi$ and $\theta$ represent two transformation functions, implemented using two 1$\times$1 convolution layers followed by a reshaping operation (\ie, $c\times h\times w \to c\times hw$).
Meanwhile, we utilize another transformation function, $\varphi$, on the projected attribute embedding $x_{proA}$ and conduct a matrix multiplication operation between it and the transpose of $S$.
Finally, we scale it by a parameter $\beta$ and conduct an element-wise summation with the image embedding $x_{I}$ to produce the attribute-image fusion feature $x_{AI}\in \mathbb{R}^{c\times h\times w}$:
\begin{equation}
x_{AI} = \beta S^{T} \varphi(x_{I})+x_{I}.
\label{attribute-image fusion feature}
\end{equation}
$x_{AI}$ serves as input to the image decoder $g_{I}$ for generating the predicted segmentation mask $P$.
In this scenario, the attribute-image fusion feature enables a comprehensive contextual understanding by merging spatial details from medical images with semantic cues from textual descriptions. This fusion equips the model to leverage complementary insights from both modalities, thus improving segmentation accuracy and efficiently capturing infected areas in medical images.

\subsection{Self-Training Mask Refinement}
\label{sec:STMR}
While the aforementioned modules leverage text attribute knowledge to assist in model training, the resulting masks may still lack precision. 
Unsupervised segmentation of lung infection regions poses challenges due to limited labeled data and the necessity to enhance segmentation accuracy. 
Thus, we introduce the self-training mask refinement process to enhance the mask by generating pseudo-labels from high-confidence predictions.
By incorporating the approach, the model iteratively enhances its segmentation performance, learning from its predictions and adjusting to data intricacies.

Specifically, we select high-probability predictions from the prediction mask $P$, filtering out low-probability ones to enable obtaining credible self-training pseudo-labels $\bar{Y} \in \mathbb{R}^{1\times H\times W}$:
\begin{equation}
\bar{Y}=\mathbbm{1}[\sigma(P)>\delta].
\label{selftraining pseudo labels}
\end{equation}
Here, $\delta$ represents a predefined threshold to discard noisy pseudo-labels. 
Subsequently, we formulate a self-training segmentation loss based on these refined pseudo-labels and prediction masks:
\begin{equation}
L_{st} = L_{seg}(P,\bar{Y}),
\label{self-training segmentation loss}
\end{equation}
$L_{seg}$ denotes the loss function commonly used in medical image segmentation, consisting of the cross-entropy loss and the Dice loss:
\begin{equation}
L_{seg} = \frac{1}{2}*L_{ce}(P,Y)+\frac{1}{2}*L_{dice}(P,Y).
\label{seg loss}
\end{equation}
$P$ indicates the prediction masks and $Y$ indicates the target labels.
Through the proposed self-training segmentation loss $L_{st}$, the model iteratively enhances its segmentation performance by updating based on high-confidence pseudo-labels, leading to more refined segmentation masks. Furthermore, this iterative refinement strategy effectively guides the model's attention to complex areas where confident predictions can be made, facilitating the convergence of segmentation masks towards improved accuracy.

\subsection{Loss Function}
\label{sec:loss}
The overall loss function for training the proposed framework contains three terms:
\begin{equation}
L_{total} = \lambda_{c}L_{c}+\lambda_{a}*L_{a}+\lambda_{st}*L_{st},
\label{overall loss}
\end{equation}
where $\lambda_{st}$, $\lambda_{a}$ and $\lambda_{st}$ are hyperparameters controlling the trade-off between different loss components.
$L_{c}$ represent a coarse segmentation loss calculated from the coarse masks in Section \ref{sec:CMG}:
\begin{equation}
L_{c} = L_{seg}(P,\hat{Y}).
\label{coarse segmentation loss}
\end{equation}
$L_{a}$ indicates the attribute classification loss defined in Eq. \ref{attribute classification loss}, and $L_{st}$ is the self-training segmentation loss  defined in Eq. \ref{self-training segmentation loss}.
During training, the weights of the attribute encoder are frozen, while the weights of the other components are updated.

\section{Experimental Results}
\label{sec:experimental}
\subsection{Experimental Settings}
\subsubsection{Datasets and Evaluation metrics.}
We evaluated our proposed framework using the QaTa-COV19 dataset \cite{degerli2022osegnet}, which consists of lung X-ray images paired with corresponding textual descriptions. 
The image data were compiled by Qatar University and Tampere University, while textual descriptions were provided by Li \etal \cite{li2023lvit}. 
Following \cite{li2023lvit}, we allocated 5716 samples for training, 1429 for validation, and 2113 for testing. 
Additionally, we rectified errors in the textual descriptions, including spelling mistakes. 
Evaluation metrics such as the Dice coefficient and Jaccard coefficient were employed to gauge the performance of our framework.

\subsubsection{Implementation Details.}
We utilize UNet \cite{ronneberger2015u}, a widely employed architecture in medical image segmentation, serving both as the image encoder and decoder. The BERT-embedding model \cite{devlin2018bert} is employed as the attribute encoder. The attribute projection employs a one-dimensional convolutional layer with a kernel size of 3. The weights of the attribute encoder are initialized with pre-trained BERT models \cite{devlin2018bert}, while those of other components are randomly initialized. The number of attribute classifiers, denoted as $M$, is set to 4. Each attribute classifier comprises two fully-connected layers with intermediate ReLU functions, and average pooling precedes the features before they are fed into the fully-connected layers. We adopt the unsupervised lung saliency detection model \cite{de2023deep}, represented as $N_{sal}$, for extracting coarse masks. The input image size is 224x224, with random rotation and flipping. We employ the Adam optimizer with a learning rate of 1e-4 and set the batch size to 12. Parameters $\tau$ and $\alpha$ are both set to 0.5, while $\delta$ is set to 0.7. Hyperparameters $\lambda_{a}$ and $\lambda_{st}$ are both set to 0.9 and 1.0, respectively.

\subsection{Comparison Results}
\begin{table*}\small
\renewcommand\arraystretch{1.1}
\caption{\label{tab:compare} 
Quantitative comparison results on the QaTa-COV19 dataset.
We report the results in terms of Dice and Jaccard.
We also report the number of model parameters Param and the computational complexity Flops.
GT refers to ground-truth masks, while CM indicates the utilization of extracted coarse masks as training targets.
}
\setlength{\tabcolsep}{0.8mm}
\begin{center}
\begin{tabular}{c|ccccc|cc}
\hline
Method&Text&Mask&Label ratio&Param (M)&Flops (G)&Dice (\%)$\uparrow$&Jaccard  (\%)$\uparrow$ \\
\hline
UNet\cite{ronneberger2015u}&$\times$&GT&100\%&14.8&50.3&79.0&69.5 \\
\hline
LAVT\cite{yang2022lavt}&$\surd$&GT&100\%&118.6&83.8&79.3&69.9 \\
\hline
LViT\cite{li2023lvit}&$\surd$&GT&100\%&29.7&54.1&83.6&75.1 \\
\hline
\hline
UNet\cite{ronneberger2015u}&$\times$&CM&0\%&14.8&50.3&45.1&32.5 \\
\hline
LViT\cite{li2023lvit}&$\surd$&CM&0\%&29.7&54.1&49.6&37.3 \\
\hline
AKGNet-T&$\surd$&CM&0\%&16.8&50.7&53.8&41.8 \\
\hline
AKGNet-I&$\surd$&CM&0\%&16.8&50.7&\textbf{55.5}&\textbf{43.7} \\
\hline
\end{tabular}
\vskip -.2in
\end{center}
\end{table*}

We initially compared AKGNet with two state-of-the-art medical image segmentation methods, UNet \cite{ronneberger2015u} and LViT \cite{yang2022lavt}, on the QaTa-COV19 dataset under the same experimental setup of image-text unsupervised segmentation. 
These methods were re-implemented and trained under identical settings as AKGNet, utilizing the generated coarse masks. 
Additionally, we conducted a comparison in a fully supervised experimental setup, involving LAVT \cite{yang2022lavt} and LViT \cite{yang2022lavt}, and excluding textual descriptions (UNet). 
The comparison results, summarized in Table \ref{tab:compare}, indicate that our proposed framework outperforms other methods in the same unsupervised scenarios, achieving a Dice value of 53.8 and a Jaccard value of 41.8 under transductive, and a Dice value of 55.5 and a Jaccard value of 43.7 under inductive settings. 
Although LViT incorporates textual descriptive information and employs a hybrid CNN and transformer structure, its results are inferior to ours, accompanied by higher parameter count and computational complexity. 
Experimental results show that our proposed AKGNet is able to achieve the best results in unsupervised experimental scenarios while taking into account less computational overhead.

\subsection{Ablation Studies}

\begin{table*}[t]\small
\renewcommand\arraystretch{1.1}
\caption{\label{tab:ablation} 
Ablation study of the proposed components on the QaTa-COV19 dataset in terms of Dice and Jaccard.
CM: generating coarse masks.
ArT: using attributes instead of text descriptions as input to the framework.
AICA: using the proposed attribute-image cross-attention module.
}
\setlength{\tabcolsep}{1.8mm}
\begin{center}
\begin{tabular}{cccccc|cc|cc}
\hline
&&&&&&\multicolumn{2}{c|}{Transductive}&\multicolumn{2}{c}{Inductive} \\
CM&$L_{c}$&$L_{a}$&ArT&AICA&$L_{st}$&Dice$\uparrow$&Jaccard$\uparrow$&Dice$\uparrow$&Jaccard$\uparrow$ \\
\hline
$\surd$&-&-&-&-&-&35.2&23.9&-&- \\
\hline
$\surd$&$\surd$&-&-&-&-&45.1&32.5&44.1&31.6 \\
\hline
$\surd$&$\surd$&$\surd$&-&-&-&47.6&34.6&48.6&36.0 \\
\hline
$\surd$&$\surd$&$\surd$&-&$\surd$&-&48.9&36.2&49.2&36.8 \\
\hline
$\surd$&$\surd$&$\surd$&$\surd$&$\surd$&-&49.7&38.8&50.3&39.4 \\
\hline
$\surd$&$\surd$&-&$\surd$&$\surd$&$\surd$&51.6&39.2&52.9&40.7 \\
\hline
$\surd$&$\surd$&$\surd$&$\surd$&$\surd$&$\surd$&53.8&41.8&55.5&43.7 \\
\hline
\end{tabular}
\vskip -.2in
\end{center}
\end{table*}

\subsubsection{Impact of different components.}

We summarize the impact of different components on the QaTa-COV19 dataset in terms of Dice value, as depicted in Table \ref{tab:ablation}.
We initially employ only the generated coarse masks as the segmentation results, yielding a Dice value of 35.2.
In the transductive scenario, the experimental results improve to a Dice value of 45.1 when utilizing the coarse masks as targets for training the segmentation model.
Further enhancements are observed when computing $L_{a}$ or incorporating the AICA module, resulting in Dice values of 47.6 and 48.9, respectively.
Moreover, performance increases to 49.7 when replacing input text descriptions with attribute descriptions.
The model achieves even better segmentation results with a Dice value of 51.6 when $L_{a}$ is excluded and $L_{st}$ is added.
Ultimately, our full model achieves the highest performance, reaching a Dice value of 53.8.
Similar trends are observed in the inductive scenario.
These findings underscore the effectiveness of the various components integrated into AKGNet.

\begin{figure}[t]
\centering
  \includegraphics[width=1\linewidth,height=35mm]{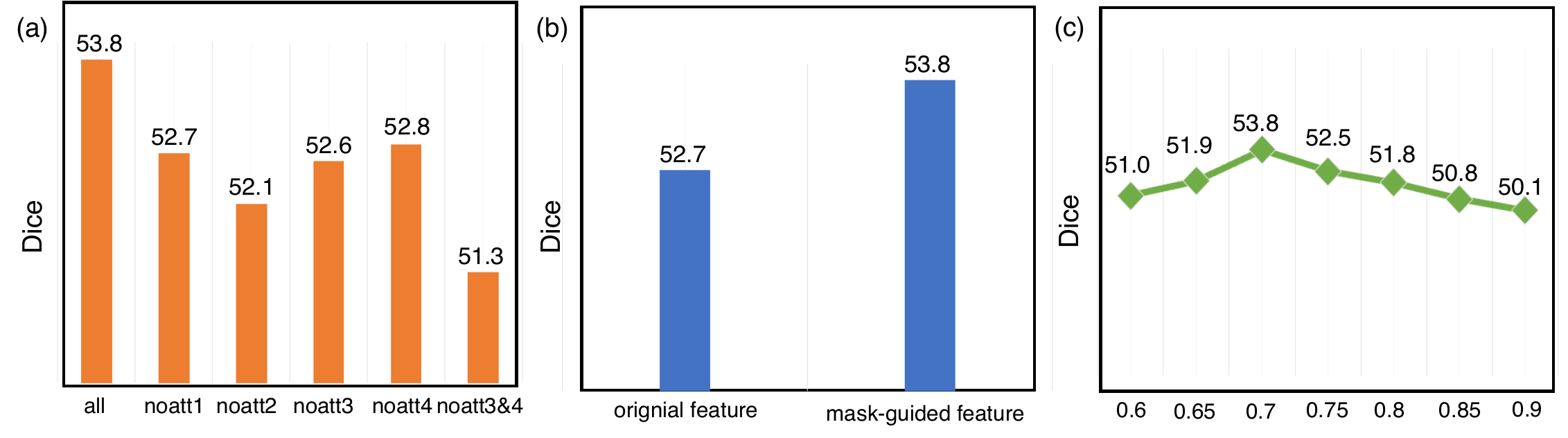}\\
\caption{
Ablation study of the 
(a) impact of different attributes in $L_{a}$;
(b) impact of using mask-guided intermediate features or original intermediate features;
(c) impact of the threshold in self-training $\delta$.
We report the Dice values.
}
  \label{fig:ablation}
\end{figure}

\subsubsection{Impact of different attributes in $L_{a}$.}
We summarize the impact of different attributes in $L_{a}$ on the QaTa-COV19 dataset for the transductive scenario in terms of Dice value, as illustrated in Fig. \ref{fig:ablation} (a).
The experimental results demonstrate that considering all attributes yields the best performance, achieving a Dice value of 53.8.
However, removing the classification loss for individual attributes leads the model to disregard crucial statistical information during training, resulting in decreased segmentation accuracy.
When omitting the first attribute (double/unilateral) or the second attribute (number), the Dice values decrease from 53.8 to 52.7 and 52.1, respectively.
Similarly, ignoring the left or right position information (third and fourth attributes) also results in performance reduction.
Furthermore, when both left and right position information are ignored, the model's performance further decreases to a Dice value of 51.3.
These experiments underscore the importance of simultaneously computing the classification loss for multiple attributes to facilitate the model in learning statistical information about different attributes.

\subsubsection{Impact of using mask-guided intermediate features or original intermediate features.}
We summarize the impact of using mask-guided intermediate features or original intermediate features as the attribute classifier's inputs on the QaTa-COV19 dataset for the transductive scenario in terms of Dice value, as depicted in Fig. \ref{fig:ablation} (b).
Utilizing the mask-guided features yields significantly better results than using the original features, with an improvement of 1.1 in terms of Dice value.
This improvement can be attributed to the fact that using mask-guided features is directly relevant to the task, and employing the prediction mask as guidance enables both the image encoder and image decoder to be trained effectively, allowing the model to generate better prediction masks while learning attribute statistics.
Importantly, our proposed architecture is capable of achieving satisfactory results using either of these two strategies, underscoring the effectiveness of our proposed method that does not rely solely on features guided by prediction masks.

\subsubsection{Impact of the threshold in self-training $\delta$.}
We summarize the impact of the threshold in self-training, denoted as $\delta$, on the QaTa-COV19 dataset for the transductive scenario in terms of Dice value, as depicted in Fig. \ref{fig:ablation} (c).
This parameter governs the confidence level in the self-training process to obtain high-confidence pseudo-labels, where larger values indicate that masks are filtered through higher confidence to obtain pseudo-labels, and vice versa.
Experimental results reveal that the optimal performance is achieved when the value is set to 0.7, yielding a Dice value of 53.8.
Deviation from this value, either by increasing or decreasing it, leads to a decrease in experimental results.
This is because excessively high confidence levels result in too few pseudo-labels, while overly low confidence levels introduce too much noise in the pseudo-labels, both of which are detrimental to achieving optimal results through self-training.
\begin{figure}[t]
\centering
  \includegraphics[width=1\linewidth,height=35mm]{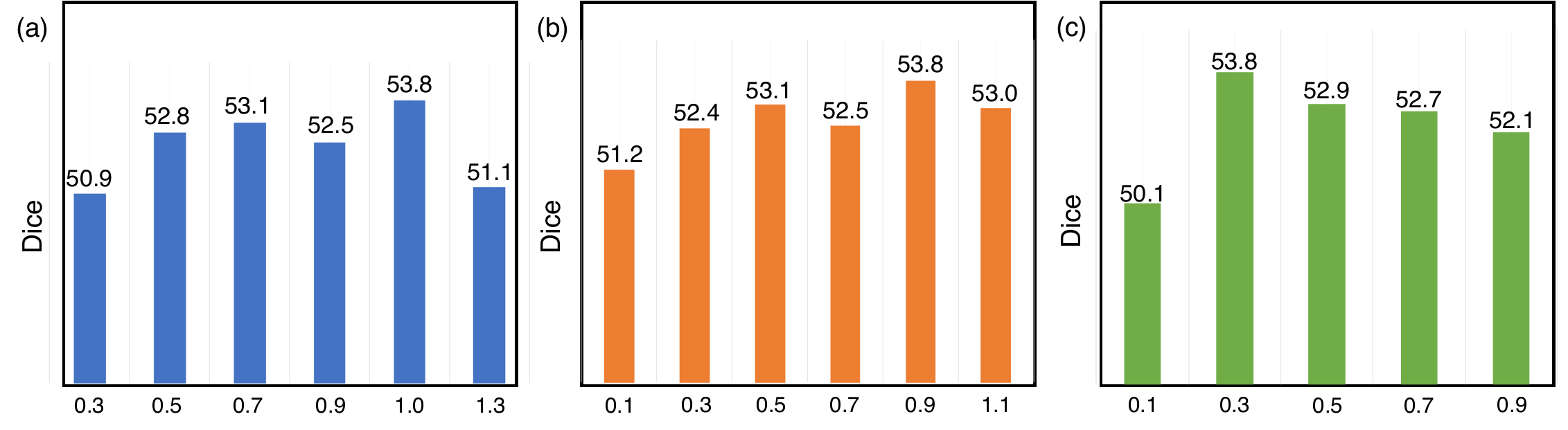}\\
\caption{
Impact of the weight of different losses
(a) $\lambda_{c}$;
(b) $\lambda_{a}$ and
(c) $\lambda_{st}$.
We report the Dice values.
}
  \label{fig:weight of losses}
\end{figure}
\subsubsection{Impact of the weight of different losses.}
We summarize the impact of the weight of different losses on the QaTa-COV19 dataset for the transductive scenario in terms of Dice value, as illustrated in Fig. \ref{fig:weight of losses}.
When adjusting the weight, all other weights remain fixed.
Firstly, concerning the coarse segmentation loss $L_{c}$ (Fig. \ref{fig:weight of losses} (a)), the optimal result is attained when it is set to 1, yielding a Dice value of 53.8.
As the weight is reduced, the results progressively deteriorate, indicating its crucial role in providing the model with a task-related training objective. Conversely, increasing its weight leads to decreased segmentation effectiveness, suggesting that excessive noise within it could have an adverse impact.
Secondly, for the attribute classification loss $L_{a}$ (Fig. \ref{fig:weight of losses} (b)), the model achieves its best results when set to 0.9. Conversely, reducing or increasing its weight may cause the model to disregard statistical information or overly focus on categorical information, thereby resulting in inferior outcomes.
Finally, concerning the self-training segmentation loss $L_{st}$ (Fig. \ref{fig:weight of losses} (c)), the optimal result is obtained when set to 0.3. Setting it to a larger value (\eg, 0.9) slightly degrades the result, suggesting that an excessively large weight during self-training could overly bias the model towards current predictions, diminishing its efficacy. Conversely, reducing it to 0.1 diminishes the extent to which the model undergoes self-training.

\subsection{Qualitative Evaluation Results}
\begin{figure}[t]
\centering
  \includegraphics[width=0.9\linewidth,height=65mm]{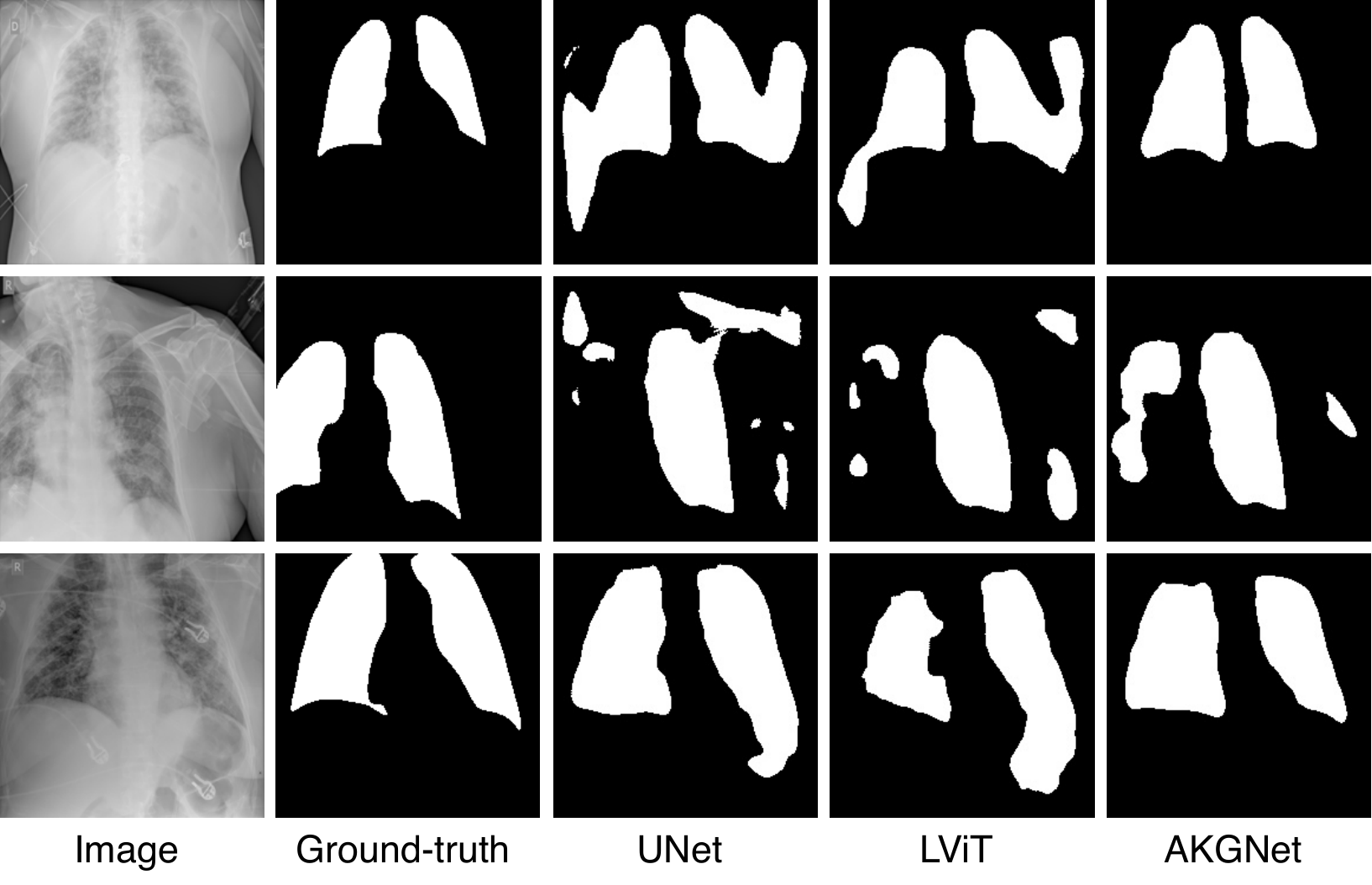}\\
\caption{
Visualization examples of different methods.
The left two columns represent input images and ground-truths.
The rightmost column represents the segmentation results of our proposed AKGNet, and the remaining two columns represent the results of UNet and LViT.
}
  \label{fig:visualization}
\end{figure}
To illustrate the effectiveness of AKGNet, we present qualitative results comparing our framework with state-of-the-art methods under unsupervised settings on the QaTa-COV19 dataset.
Visualizations reveal that AKGNet outperforms other methods, particularly in scenarios with asymmetric infected regions or when these regions are near the image edge.
UNet, when used without textual guidance, tends to produce numerous mis-segmentations.
While LViT utilizes textual descriptive information, its performance in unsupervised scenarios falls short compared to AKGNet.

\section{Conclusion}
In this paper, we introduce a novel AKGNet framework for image-text unsupervised lung-infected areas segmentation.
AKGNet uses a text attribute knowledge learning module to learn statistical information and facilitate model adaptation to various text attributes by excavating text attribute knowledge.
Additionally, an image-attribute cross-attention module is used to capture spatial dependencies between images and text attributes by computing correlations in the embedding space.
Furthermore, a self-training mask refinement process is employed to accelerate model convergence towards refined masks.
Experimental results demonstrate the effectiveness of the proposed framework, surpassing existing segmentation methods in unsupervised scenarios.

\bibliographystyle{splncs04}
\bibliography{mybibliography}
\end{document}